\documentclass[conference]{IEEEtran}
\IEEEoverridecommandlockouts
\usepackage{cite}
\usepackage{amsmath,amssymb,amsfonts}
\usepackage{algorithmic}
\usepackage{graphicx}
\usepackage{textcomp}
\usepackage{xcolor}
\def\BibTeX{{\rm B\kern-.05em{\sc i\kern-.025em b}\kern-.08em
    T\kern-.1667em\lower.7ex\hbox{E}\kern-.125emX}}

\graphicspath{{Figs/}}
\DeclareGraphicsExtensions{.pdf,.jpeg,.png,.svg,.eps}

\usepackage{url}
\ifCLASSOPTIONcompsoc
\usepackage[caption=false,font=normalsize,labelfon
t=sf,textfont=sf]{subfig}
\else
\usepackage[caption=false,font=footnotesize]{subfi
g}
\fi
\usepackage{booktabs}
\usepackage{multirow}

\newcommand{\positenv}[2]{Posit$\langle #1, #2 \rangle$}
\newcommand{\positnotation}[2]{$\langle #1, #2 \rangle$}
\newcommand{\ra}[1]{\renewcommand{\arraystretch}{#1}}

\usepackage{array}
\newcolumntype{L}{>{$}l<{$}}
\newcolumntype{C}{>{$}c<{$}}
\newcolumntype{R}{>{$}r<{$}}

\usepackage{fancyhdr}
\usepackage{kantlipsum}

\fancypagestyle{plain}{
  \fancyhf{}
  \fancyhead[C]{Conference on \LaTeX}     
  \fancyfoot[L]{This is a notice}

}
\usepackage{eso-pic}

\AddToShipoutPictureBG*{%
  \AtPageUpperLeft{%
    \setlength\unitlength{1in}%
    \hspace*{\dimexpr0.5\paperwidth\relax}
    \makebox(0,-0.5)[c]{\footnotesize This article has been accepted for publication in a future issue of this journal, but has not been fully edited. Content may change prior to final publication.}%
}}
\AddToShipoutPictureBG*{%
  \AtPageUpperLeft{%
    \setlength\unitlength{1in}%
    \hspace*{\dimexpr0.5\paperwidth\relax}
    \makebox(0,-0.8)[c]{\footnotesize Citation information: DOI 10.1109/TETC.2021.3109127, IEEE Transactions on Emerging Topics in Computing}%
}}

\begin{document}
\bstctlcite{IEEEexample:BSTcontrol}
%
\title{PLAM: A Posit Logarithm-Approximate Multiplier}

\author{\IEEEauthorblockN{Raul Murillo\IEEEauthorrefmark{1},
Alberto A. Del Barrio\IEEEauthorrefmark{1},
Guillermo Botella\IEEEauthorrefmark{1},
Min Soo Kim\IEEEauthorrefmark{2},
HyunJin Kim\IEEEauthorrefmark{3},
Nader Bagherzadeh\IEEEauthorrefmark{4}}
\IEEEauthorblockA{\IEEEauthorrefmark{1}Department of Computer Architecture and Automation, 
Complutense University of Madrid, 28040 Madrid, Spain}
\IEEEauthorblockA{\IEEEauthorrefmark{2}NGD Systems, Irvine, California 92618, USA}
\IEEEauthorblockA{\IEEEauthorrefmark{3}School of Electronics and Electrical Engineering, Dankook University, Yongin-si, Gyeonggi-do 16890, Republic of Korea}
\IEEEauthorblockA{\IEEEauthorrefmark{4}Department of Electrical Engineering and Computer Science, University of California, Irvine, California 92697, USA}
\IEEEauthorblockA{Email: \IEEEauthorrefmark{1}\{ramuri01, abarriog, gbotella\}@ucm.es, \IEEEauthorrefmark{2}minsk1@uci.edu, \IEEEauthorrefmark{3}hyunjin2.kim@gmail.com, \IEEEauthorrefmark{4}nader@uci.edu}
}

%

\maketitle

\begin{abstract}
The Posit™ Number System was introduced in 2017 as a replacement for floating-point numbers. Since then, the community has explored its application in Neural Network related tasks and produced some unit designs which are still far from being competitive with their floating-point counterparts. This paper proposes a Posit Logarithm-Approximate Multiplication (PLAM) scheme to significantly reduce the complexity of posit multipliers, the most power-hungry units within Deep Neural Network architectures. When comparing with state-of-the-art posit multipliers, experiments show that the proposed technique reduces the area, power, and delay of hardware multipliers up to 72.86\%, 81.79\%, and 17.01\%, respectively, without accuracy degradation.

\end{abstract}


\section{Introduction}
\label{sec:introduction}

Deep Neural Networks (DNNs) dominate nowadays machine learning landscape due to the great advances in a large variety of applications, including computer vision, natural language processing or speech recognition. This continuous improvement of the state of the art has been accompanied by an increase in computational complexity and an overhead in hardware resources. While DNN models are commonly trained on high-end GPUs, reducing the computational complexity of DNNs to perform inference on resource-constraint devices has been a serious challenge and a lengthy line of research \cite{Bianco2018}.

The IEEE 754 standard for floating-point arithmetic has been for decades the de facto implementation for the vast majority of real number-based applications. However, in the last years different computer arithmetic encodings and formats have been considered for DNN training and inference, including half-precision, fixed-point, bfloat16, or 8-bit integer quantization \cite{Hennessy2019}. But probably, one of the most promising alternatives to the floating-point standard is the Posit Number System (PNS) \cite{Gustafson2017a}. Posit numbers provide a better trade-off than floating point between dynamic range and numerical precision, with a larger dynamic range under the same bitwidth
and, most importantly, tapered accuracy around $\pm 1$, corresponding with the DNN weight distribution \cite{Murillo2020}.

\begin{figure}[!t]
\centering
\subfloat[Area]{\includegraphics[width=0.49\linewidth]{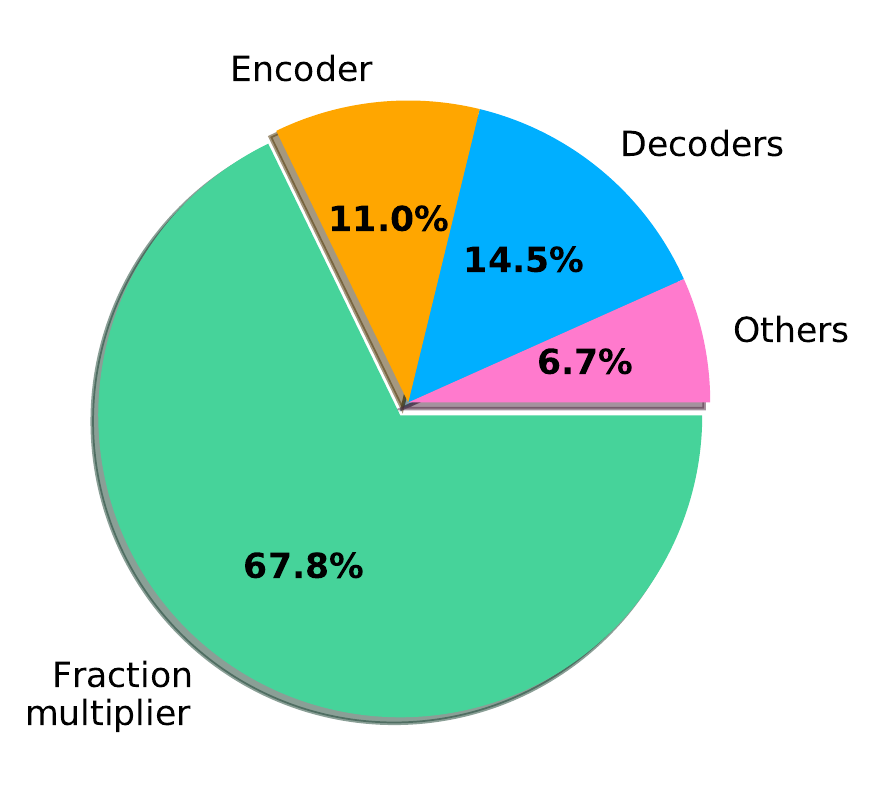}%
\label{fig:area_perc}}
\hfil
\subfloat[Power consumption]{\includegraphics[width=0.51\linewidth]{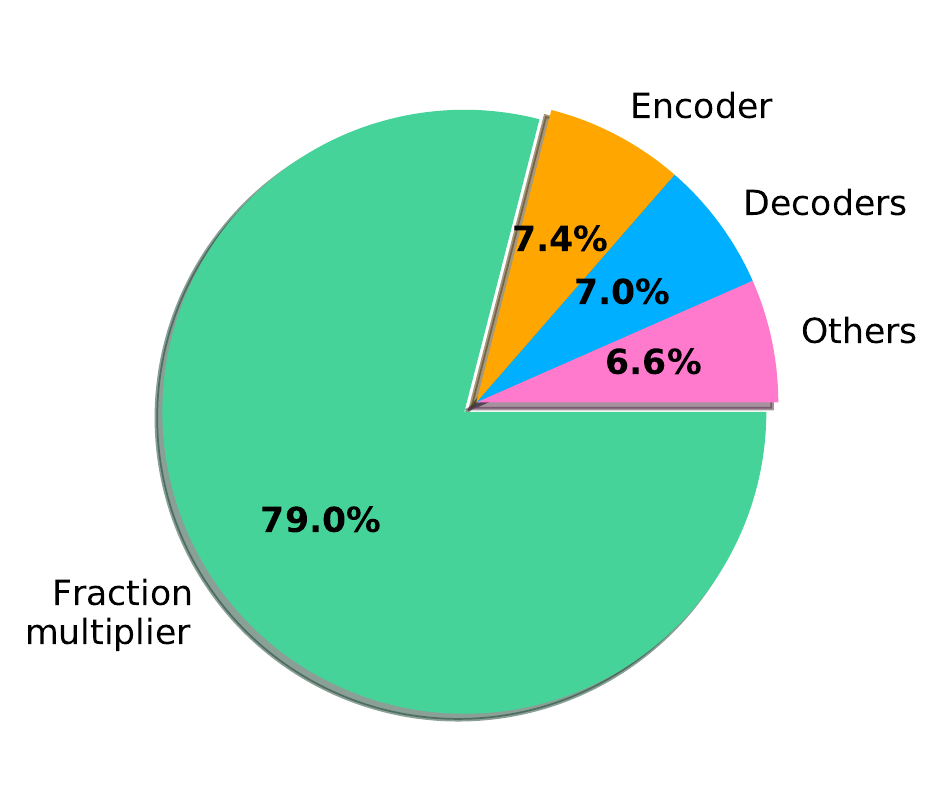}%
\label{fig:power_perc}}
\caption{Resource distribution of a \positenv{32}{2} multiplier.}
\label{fig:percent}
\end{figure}

One of the most widespread techniques in DNNs is quantization \cite{Krishnamoorthi2018}, which allows the bitwidth reduction of the units deployed in the corresponding circuit. This contributes diminishing area, power or memory footprint. Furthermore, approximate computing techniques \cite{Lotric2012}
have commonly been used in DNNs too, specially for inference in real time systems and resource-constraint devices, where the trade off between inference accuracy and execution time or resources is always present.
The inference stage mainly consists of addition and multiplication operations, the latter being the most expensive one from a hardware perspective.
Despite the fact that the PNS has shown its potential in DNNs \cite{Johnson2018, Carmichael2019a, Langroudi2019b, Murillo2020, Lu2020}, it is true that posit units are far from being competitive in terms of power with respect to FP or bfloat16 formats \cite{Jaiswal2018a, Jaiswal2018, Chaurasiya2019, Jaiswal2019, Uguen2019,  murillo2020customized}.
Without much details, which will be commented in Section \ref{sec:logarithm_approx}, a posit multiplier consists of similar stages to a IEEE 754 floating-point multiplier, i.e. unpacking/decoding of operands, fraction multiplication and packing/encoding of the result. As can be seen in Fig. \ref{fig:percent}, the fraction multiplier is, by far, the module with the highest consumption of resources.
For this reason, as in the case of floating-point formats, reducing the complexity of the fraction multiplier is critical to optimize the power consumption of the whole unit.

To improve the efficiency of posit-based DNN inference, this work proposes the use of logarithm-approximate multipliers in combination with the PNS.
Experimental results reveal that adopting the proposed Posit Logarithm-Approximate Multiplies (PLAM) allows to significantly reduce area, power, and delay with negligible accuracy degradation while taking advantage of the benefits of this novel arithmetic format.
To the best of our knowledge, this is the first work proposing the use of approximate posit multipliers.
Thus, the main contributions of this paper can be summarized as follows:
\begin{itemize}
    \item Proposing an algorithm for performing logarithm-approximate multiplication in posit arithmetic, with a bounded error of $11.1\%$.
    \item Testing the proposed algorithm at inference for several DNNs including LeNet-5 and CifarNet, and well-known datasets as MNIST, SVHN and CIFAR-10, achieving negligible accuracy degradation with 16-bit posits.
    \item Implementing the proposed algorithm in the open-source FloPoCo framework, reducing area and power up to $72.86\%$ and $81.79\%$, respectively, when comparing to cutting edge posit implementations.
\end{itemize}

The rest of the paper is organized as follows:
Section \ref{sec:related work} reviews strategies for using posits in NNs and implementations from related works.
Section \ref{sec:logarithm_approx} introduces PLAM and discusses its approximation error.
Experimental results of adopting PLAM for inference in different DNNs are presented in Section \ref{sec:experiments}.
Section \ref{sec:hw_eval} presents evaluations of area, power, and  delay, and discusses the improvements of PLAM against previous posit and floating-point implementations.
Finally, Section \ref{sec:conclusions} concludes this paper.

\section{Related Work}
\label{sec:related work}

\subsection{The Posit Number System}

\begin{figure}[!t]
    \centering
    \includegraphics[width=\linewidth]{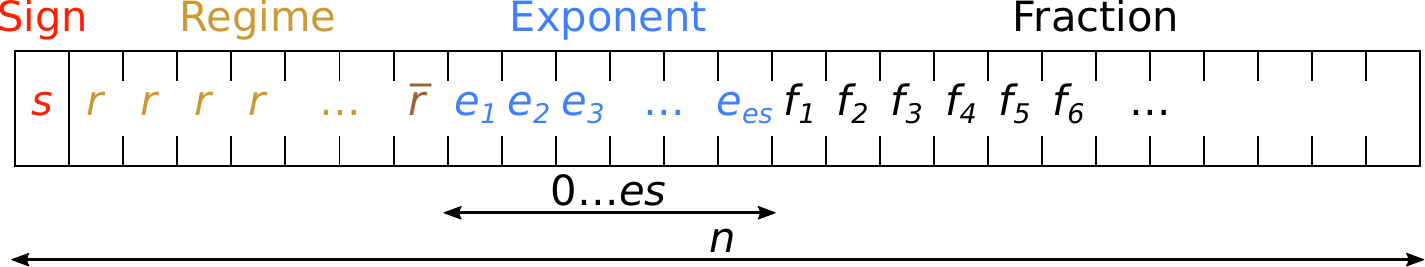}
    \caption{Layout of an \positenv{n}{es} number. The variable-length regime field may cause exponent be encoded with less than $es$ bits, even no bits if regime is wide enough. Same occurs with the fraction.}
    \label{fig:posit_format}
\end{figure}

A posit format is defined as a tuple \positnotation{n}{es}, where $n$ is the total bitwidth and $es$ is the maximum number of bits reserved for the exponent field. 
As Fig. \ref{fig:posit_format} shows, posit numbers are encoded with four fields:
a sign bit ($s$), several bits for encoding the regime value ($k$), up to $es$ bits for the exponent ($e$), and the remaining bits for fraction ($f$). Thus, the numerical value $X$ of a generic \positenv{n}{es} is expressed by \eqref{eq:posit_decimal_value}.

\begin{align}
    \label{eq:posit_decimal_value}
    X & = (-1)^{s}\times (2^{2^{es}})^{k} \times 2^{e} \times (1 + f),\\
    \label{eq:posit_regime}
    k & = - x_{n-2} + \sum_{i=n-2}^{x_{i} \neq x_{n-2}}(-1)^{1 - x_{i}}.
\end{align}

The main differences with floating-point format are the utilization of an unsigned and unbiased exponent, if there exists such exponent field, and the existence of the regime field. This new field consists of a sequence of bits with the same value finished with the negation of such value, as shown in Fig. \ref{fig:posit_format}. Provided that $X=x_{n-1}x_{n-2}...x_{1}x_{0}$, this regime can be expressed as \eqref{eq:posit_regime} shows.
It is noteworthy that, while the new regime field provides important scaling capabilities that improve the dynamic range of posits, detecting the resulting varying-sized fields adds a hardware overhead.

\subsection{The use of Posit Arithmetic in NNs}

Posits were introduced by John Gustafson in 2017. Since then, multiple works have explored the benefits of this novel format against the standard floating-point, and most of them focusing on NNs. In \cite{Johnson2018}, J. Johnson designed an arithmetic unit for combining posit addition together with logarithmic multiplication for performing CNN inferences.
Authors in \cite{Carmichael2019a} employ a posit DNN accelerator to represent weights and activations combined with an FPGA soft core for 8-bit posit exact-multiply-and-accumulate (EMAC) operations.
In all these works, DNN training is performed in floating-point, while the inference stage is performed in low-precision posit format.
Later works proposed different approaches for training NNs using the posit format, either directly training on this format with different precision \cite{Langroudi2019b, Murillo2020, Cococcioni2020a}, or with the help of a warmup training using floating-point format \cite{Lu2020}.

Several previous works have also proposed to introduce approximate multipliers into DNN inferences. 
Logarithmic multiplication has successfully been employed to fixed-point small DNN models in \cite{Kim_2019} and even to large ones in \cite{Kim2020}. Finally, authors in \cite{Cheng2020} explore its use within floating-point multipliers to reduce the costs of small NNs during the training and inference stages.
To the best of our knowledge, at the time of writing there are no previous works proposing the design of approximate posit multipliers and its use for deep learning tasks.
With regard to other approximate computing techniques on posit arithmetic, Cococcioni et al. \cite{Cococcioni2020a} explored the possibility to approximate simple operations and activation functions in DNNs using only ALU-based operations.

\subsection{Posit Arithmetic-Based Implementations}

Since the appearance of the PNS, several hardware implementations for this arithmetic format have been proposed. An open-source parameterized adder/subtractor was presented in \cite{Jaiswal2018a}, whose concepts where expanded in \cite{Jaiswal2018} to design a parameterized posit multiplier. These two works did not perform posit rounding, but fraction truncation, and used both a Leading One Detector (LOD) and a Leading Zero Detector (LZD) to determine regime value, which results in redundant area. This shortcomings were solved in \cite{Chaurasiya2019}, where only a LZD was used at the cost of inverting negative regimes, and results were correctly rounded using the round to nearest even scheme. The same idea was applied in \cite{Jaiswal2019}, where authors expanded their previous works \cite{Jaiswal2018a, Jaiswal2018} and presented an open-source posit core generator which included a parameterized divider based on the Newton-Raphson method. Another open-source posit core generator is presented \cite{murillo2020customized}, where parameterized adder and multiplier designs were integrated into the FloPoCo framework, allowing even to generate posit units with no exponent bits, in contrast with previous works.
However, posits are still in development. As has been mentioned, in terms of delay, area, and power, the arithmetic units are not yet competitive against their floating-point counterparts, and although they have shown some promising improvements in the NNs field \cite{Murillo2020, Langroudi2019b} there is still some debate about their real improvement \cite{DeDinechin2019}.

In this work, we propose the design of a log-based approximate posit multiplier to perform DNN inference without degrading the accuracy of the results.
With this approach, it is possible to drastically reduce the complexity of the posit multiplication.

\section{PLAM: the posit logarithm-approximate multiplier}
\label{sec:logarithm_approx}

Hardware multiplication is an expensive operation, specially between two real numbers, where all the state-of-the-art formats mandate multiplying two fixed-point values. As depicted in Fig. \ref{fig:percent} for the posit format, this multiplication provokes the majority of the power consumption.
Instead, logarithm multiplication \cite{Mitchell1962, Kim_2019}
avoids hardware multipliers entirely by approximating the multiplication as a fixed-point addition. This section introduces the Posit Logarithm-Approximate Multiplier (PLAM), and analyzes its approximation error.

\subsection{Exact posit multiplication}

\begin{figure}[!t]
    \centering
    \includegraphics[width=\linewidth]{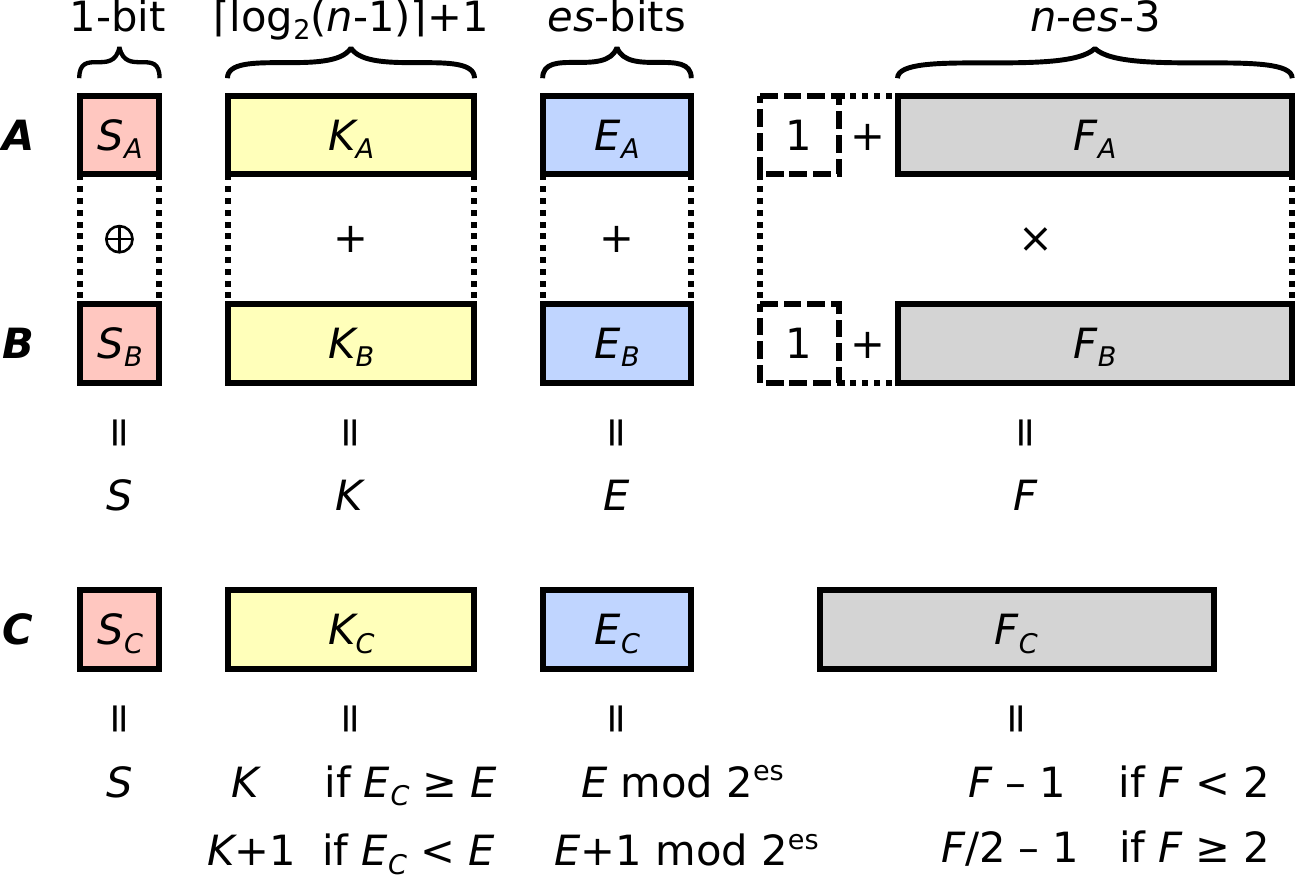}
    \caption{Exact posit multiplication.}
    \label{fig:posit_mult}
\end{figure}

As has been mentioned, while posit encoding may differ from usual floating-point, the core of the operations is quite similar between these number formats, with exception on the decoding and encoding of the posit fields \cite{murillo2020customized}.
In addition to this, in the PNS there are no special cases to being taken care of, as the denormal numbers in the case of floating-point based formats, a single rounding mode, i.e. round to nearest even, and unique representations for zero and infinite values.
Provided that a posit number $X$ is represented by the tuple $(S_X, K_X, E_X, F_X)$, where $S_X$, $K_X$, $E_X$, $F_X$, are the sign, regime, exponent and fraction values, respectively, the multiplication of two posit values $C = A \times B$ is depicted in Fig. \ref{fig:posit_mult}.
The computation of the different fields is defined by \eqref{eq:posit_mult_sign} to \eqref{eq:posit_mult_frac}.
\begin{align}
    \label{eq:posit_mult_sign}
    S & = S_A \oplus S_B,\\
    \label{eq:posit_mult_regime}
    K & = K_A + K_B,\\
    \label{eq:posit_mult_exp}
    E & = E_A + E_B,\\
    \label{eq:posit_mult_frac}
    F & = (1 + F_A) \times (1 + F_B).
\end{align}
Note that $(1 + F_A)$ (respectively $1 + F_B$) is obtained by appending a hidden bit with value 1 to the binary representation of $F_A$ (respectively $F_B$). Therefore, the resulting posit $C$ is obtained as described in  \eqref{eq:posit_mult_final_sign} to \eqref{eq:posit_mult_final_frac}.
\begin{align}
    \label{eq:posit_mult_final_sign}
    S_C & = S,\\
    \label{eq:posit_mult_final_regime}
    K_C & = \begin{cases}
                K &\text{if }E_C \ge E,\\
                K + 1 & \text{otherwise},
            \end{cases}\\
    \label{eq:posit_mult_final_exp}
    E_C & = \begin{cases}
                E \mod 2^{es} & \text{if } F < 2,
                \\
                (E + 1) \mod  2^{es} & \text{otherwise},
            \end{cases}\\
    \label{eq:posit_mult_final_frac}
    F_C & = \begin{cases}
                F - 1 &\text{if }F < 2,\\
                F/2 - 1 &\text{otherwise}.
            \end{cases}
\end{align}

\subsection{Logarithm approximated posit multiplication}

Once the posit multiplication is explained, the insights on how to approximate this operation taking advantage of the logarithmic multiplication will be described in detail.

Firstly, it is worthy to note that in the multiplication operation the computation of the sign is independent to the computation of the other fields. Therefore, let us focus on positive posit numbers from now on.
In such case, \eqref{eq:posit_decimal_value} is simplified as shown in \eqref{eq:posit_positive},
\begin{equation}
    \label{eq:posit_positive}
    X = (2^{2^{es}})^{k} \times 2^{e} \times (1 + f).
\end{equation}
Taking logarithms on both sides, \eqref{eq:posit_positive} becomes
\begin{equation}
    \label{eq:posit_log}
    \log_{2} X = 2^{es} \times k + e + \log_{2}(1 + f) \approx 2^{es} \times k + e + f,
\end{equation}
where the approximation of the right term is based on the property
\begin{equation}
    \label{eq:log_approx}
    \log_{2} (1 + x) \approx x, \text{ for } 0 \leq x \leq 1.
\end{equation}

Converting numbers to the logarithmic domain allows computing the multiplication as the addition of fixed point numbers. Thus, in order to compute the multiplication of two posits $C = A \times B$ as a logarithm-approximate multiplication, the different fields are processed as described in \eqref{eq:posit_lam_sign} to \eqref{eq:posit_lam_frac}.
\begin{align}
    \label{eq:posit_lam_sign}
    S & = S_A \oplus S_B,\\
    \label{eq:posit_lam_regime}
    K & = K_A + K_B,\\
    \label{eq:posit_lam_exp}
    E & = E_A + E_B,\\
    \label{eq:posit_lam_frac}
    F & = F_A + F_B,
\end{align}
where \eqref{eq:posit_lam_sign}, \eqref{eq:posit_lam_regime} and \eqref{eq:posit_lam_exp} are identical to \eqref{eq:posit_mult_sign}, \eqref{eq:posit_mult_regime} and \eqref{eq:posit_mult_exp}. The only difference is in the computation of the fraction field, where \eqref{eq:posit_lam_frac} substitutes the product of \eqref{eq:posit_mult_frac} by an addition thanks to the log property shown in \eqref{eq:log_approx}. Then, the multiplication result is expressed by \eqref{eq:posit_lam_final_sign} to \eqref{eq:posit_lam_final_frac}.
\begin{align}
    \label{eq:posit_lam_final_sign}
    S_C & = S,\\
    \label{eq:posit_lam_final_regime}
    K_C & = \begin{cases}
                K &\text{if }E_C \ge E,\\
                K + 1 & \text{otherwise},
            \end{cases}\\
    \label{eq:posit_lam_final_exp}
    E_C & = \begin{cases}
                E \mod 2^{es} & \text{if } F < 1,
                \\
                (E + 1)\mod  2^{es} & \text{otherwise},
            \end{cases}\\
    \label{eq:posit_lam_final_frac}
    F_C & = \begin{cases}
                F &\text{if }F < 1,\\
                F - 1 &\text{otherwise}.
            \end{cases}
\end{align}

\begin{figure}[!t]
    \centering
    \includegraphics[width=0.8\linewidth]{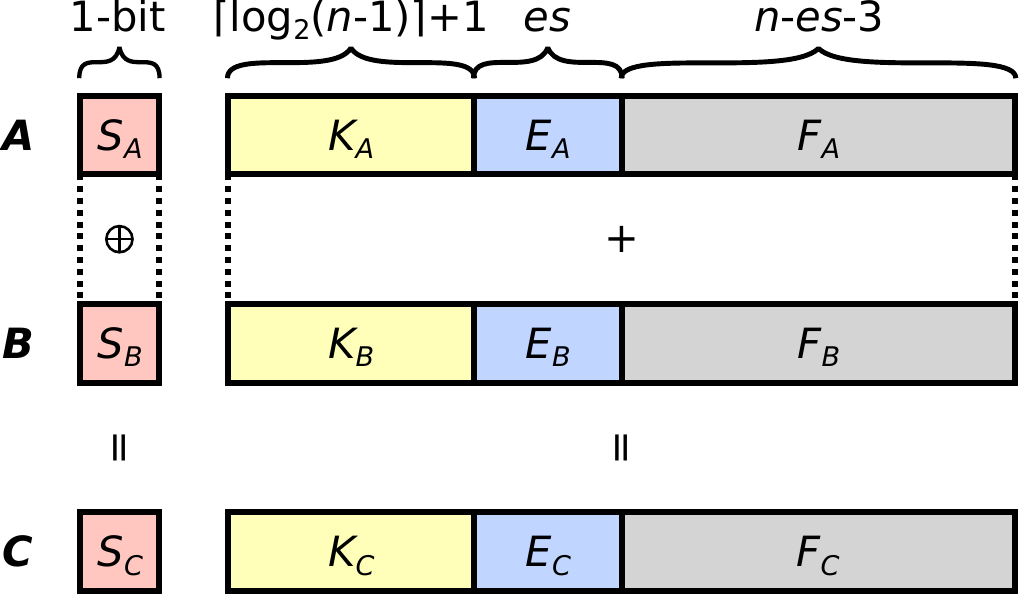}
    \caption{Algorithm for implementing PLAM in hardware.}
    \label{fig:posit_lam}
\end{figure}

Finally, let us focus on the hardware implementation of PLAM based on the algorithm presented in \eqref{eq:posit_lam_sign} to \eqref{eq:posit_lam_final_frac}.
According to \eqref{eq:posit_log}, a posit in log-domain gets its regime value $k$ multiplied by $2^{es}$ and then adds the exponent value. When implementing this in hardware, it is equivalent to concatenate both the regime and exponent bit fields. 
In this way, the condition in \eqref{eq:posit_lam_final_regime} can be efficiently computed in hardware, as the overflow coming from $E_A + E_B$ is directly added as a carry-in to $K_A + K_B$. Besides, conditions in \eqref{eq:posit_lam_final_exp} and \eqref{eq:posit_lam_final_frac} can be implemented in hardware in the same manner, as illustrated in Fig. \ref{fig:posit_lam}.

\subsection{Approximation error}

To analyze the error of PLAM, which directly depends on the input operands, let us consider two positive posit numbers $A = s_A (1 + f_A)$ and $B = s_B (1 + f_B)$, where the scaling factor $s_i = 2^{(2^{es} \times k_i + e_i)}$. In such case, the results of exact and approximate multiplication $C = A \times B$ are given by \eqref{eq:posit_exact_mult} and \eqref{eq:posit_lam}, respectively.


\begin{align}
    \label{eq:posit_exact_mult}
    C_{exact} & = s_A s_B (1 + f_A) (1 + f_B)\\
    \label{eq:posit_lam}
    C_{PLAM} & = \begin{cases}
                    s_A s_B (1 + f_A + f_B) &\text{if }f_A + f_B < 1,\\
                    2 s_A s_B (f_A + f_B) &\text{otherwise}.
                \end{cases}
\end{align}



The relative error of approximation, defined by \eqref{eq:error}, is a function of just $f_A$ and $f_B$. As these parameters are restricted to the interval $[0, 1)$, the maximum error is $11.1\%$, which is obtained when both fractions are equal to 0.5, as Mitchell demonstrated in \cite{Mitchell1962}. It is noteworthy that neither the exponents nor the novel regime fields affect the error value, it just depends on the fractions.

\begin{equation}
    \label{eq:error}
    \begin{split}
    error & = \frac{C_{exact} - C_{PLAM}}{C_{exact}} \\
     & =
    \begin{cases}
        \frac{f_A f_B}{(1 + f_A) (1 + f_B)} &\text{if }f_A + f_B < 1,\\
        \frac{(1 - f_A) (1 - f_B)}{(1 + f_A) (1 + f_B)}  &\text{otherwise}.
    \end{cases}
    \end{split}
\end{equation}

The impact of PLAM error on the DNN inference will be investigated in Section \ref{sec:experiments}, and its advantage in terms of the speed, power, and area will be discussed in Section \ref{sec:hw_eval}.


\section{Experimental results for DNN inference}
\label{sec:experiments}

This section demonstrates the effect of PLAM and posits in DNN inference.

\subsection{Experimental setup}

To demonstrate the effect of the proposed approximate posit multiplication in DNN inference, we evaluate accuracy results on different datasets and topologies.
Table \ref{table:NN_setup} lists the different datasets, network architectures and training configurations used for the experiments in this paper. For numeric datasets, fully connected networks with 2 hidden layers are employed (for which the number of neurons per layer is indicated in the table), while for image datasets, convolutional models as LeNet-5 \cite{LeCun1998} or CifarNet \cite{krizhevsky2009learning} are more appropriate. For all setups, ReLU is used as activation function of hidden layers, and softmax function is applied to the output layer.

\begin{table}[!t]\centering
    \ra{1.2}
    \caption{DNNs setup
    }
    \label{table:NN_setup}
    \begin{tabular}{@{}lllcc@{}}
    \toprule
    Dataset  & Architecture & Optimizer & Batch size & Epochs \\ \midrule
    ISOLET   & (617, 128, 64, 26) & SGD      & 64   & 30 \\
    UCI HAR  & (561, 512, 512, 6) & Nesterov & 32   & 30 \\
    MNIST    & LeNet-5            & Adam     & 128  & 50 \\
    SVHN     & LeNet-5            & Adam     & 128  & 50 \\
    CIFAR-10 & CifarNet           & Adam     & 128  & 30 \\ \bottomrule
    \end{tabular}
\end{table}

Due to the lack of hardware support for the posit number system, computations on this format are performed via software emulation.
The open-source framework Deep PeNSieve \cite{Murillo2020} allows to generate DNN models and perform inference and training entirely using posits.
Deep PeNSieve relies on the reference library SoftPosit. In this work, both libraries are extended to support PLAM operation for both scalar and matrix multiplication using the algorithm explained in Section \ref{sec:logarithm_approx}.
It must be noted that larger DNNs cannot be efficiently trained using this framework, since the lack of native support makes the training times too long. For instance, training CifarNet on an Intel\textsuperscript{®} Core™ i7-9700K processor with 32 GB of RAM takes around 10 days.

\subsection{Evaluation for DNNs}

As previous works demonstrate \cite{Murillo2020, Lu2020}, 16-bit posits can be used for DNN training with no accuracy loss with respect the baseline 32-bit floating-point format.
Accordingly, each model is trained under single-precision floating-point and \positenv{16}{1} formats. Trained posit models are then modified to use PLAM at inference stage. Three different models are trained for each dataset, and the averages of the results are presented in Table \ref{table:DNN_results}.
Approximate posit multiplication reaches similar accuracy on the inference stage as exact posit multiplication.
16-bit floating-point inference presented similar results as using single-precision, so it is omitted.
It is quite remarkable that posits perform better than floats in CIFAR-10, but as authors mention in \cite{Murillo2020}, this still need further evaluations in order to test larger architectures with regularization layers. In any case, the purpose of this paper is to prove that PLAM can perform as accurate as exact posit formats in posit-based DNNs.

\begin{table} [!t]
\centering
\ra{1.2}
\caption{Accuracy results for the inference stage}
\label{table:DNN_results}
\resizebox{\linewidth}{!}{
    \begin{tabular}{LCRCRCR}
    \toprule &
    \multicolumn{2}{c}{Float 32-bit}    &
    \multicolumn{2}{c}{\positenv{16}{1}}    &
    \multicolumn{2}{c}{\positenv{16}{1}\textsubscript{PLAM}}    \\ 
    \cmidrule(lr){2-3}
    \cmidrule(lr){4-5}
    \cmidrule(lr){6-7}

    \multicolumn{1}{l}{Dataset} &
    \multicolumn{1}{c}{Top-1}   &
    \multicolumn{1}{c}{Top-5}   &
    \multicolumn{1}{c}{Top-1}   &
    \multicolumn{1}{c}{Top-5}   &
    \multicolumn{1}{c}{Top-1}   &
    \multicolumn{1}{c}{Top-5}   \\
    \midrule
    
    \multicolumn{1}{l}{ISOLET}
        & 0.9066 & 0.9568 &  0.9093 & 0.9585 & 0.9051 & 0.9585 \\
    \multicolumn{1}{l}{UCI HAR}
        & 0.9383 & 0.9841 & 0.9307 & 0.9841 & 0.9282 & 0.9841 \\
    \multicolumn{1}{l}{MNIST}
        & 0.9907 & 0.9999 & 0.9903 & 1.0    & 0.9898 & 1.0    \\
    \multicolumn{1}{l}{SVHN}
        & 0.8624 & 0.9794 & 0.8513 & 0.9766 & 0.8489 & 0.9761 \\
    \multicolumn{1}{l}{CIFAR-10}
        & 0.6933 & 0.9722 & 0.7247 & 0.9744 & 0.7251 & 0.9743 \\
    \bottomrule
    \end{tabular}
}
\end{table}

\section{Evaluation of hardware implementation}
\label{sec:hw_eval}

After evaluating the accuracy of PLAM when performing DNN inference, this section presents synthesis results for the proposed PLAM, demonstrating the advantage of approximate units from the hardware perspective.

The proposed architecture has been implemented into FloPoCo, an open-source C++ framework for the generation of arithmetic datapaths that provides a command-line interface that inputs operator specifications and outputs synthesizable VHDL \cite{DeDinechin2011}. This tool allows operators to be automatically generated with the specified parameters and, therefore, to obtain posit operators for arbitrary values of \positnotation{n}{es} with the same base design. The proposed PLAM design, which includes support for correct rounding, has been made publicly available in the FloPoCo git repository. Simulations with extensive testing vectors are performed to verify the functionality of the proposed design. The testing vectors are generated by extending SoftPosit library with support for logarithm-approximate multiplication.

To evaluate the impact of PLAM in terms of hardware resources, 16-bit and 32-bit models (without pipelining) of the proposed architecture have been generated. These operators will be then compared with state-of-the-art implementations of posit and floating-point exact multipliers.
The latter ones have been generated using the FloPoCo library. However, it is important to mention that the designs provided by this tool do not include support for denormal numbers or full exception handling, and thus, use less resources compared with a fully IEEE-754 compliant implementation.
To have a fair comparison with the works presented in \cite{Chaurasiya2019, Uguen2019}, the proposed operators, as well as the open-source implementations (Posit-HDL \cite{Jaiswal2018}\footnote{source code accessed on November 1, 2020 from \url{https://github.com/manish-kj/Posit-HDL-Arithmetic}}, PACoGen \cite{Jaiswal2019}\footnote{source code accessed on November 1, 2020 from \url{https://github.com/manish-kj/PACoGen}} and FloPoCo-Posit \cite{murillo2020customized}\footnote{source code accessed on November 1, 2020 from \url{https://gitlab.inria.fr/fdupont/flopoco}}) have been synthesized on a Zedboard with a Zynq-7000 SoC, using Vivado 2020.1 with default settings. FPGA synthesis results, reported in Table \ref{table:FPGA_synth}, show a clear reduction in resource utilization.
It must be noted that PLAM uses less LUTs that the rest of posit multipliers as well as no DSPs.


\begin{table}[!t]\centering
    \ra{1.2}
    \caption{FPGA resource utilization}
    \label{table:FPGA_synth}
    \begin{tabular}{@{}crrcrr@{}}\toprule
        & \multicolumn{2}{c}{16-bits} & \phantom{i} & \multicolumn{2}{c}{32-bits} \\
        \cmidrule{2-3} \cmidrule{5-6}
        Work  & LUTs  & DSP  && LUTs  & DSP  \\
        \midrule
        \cite{Jaiswal2018} & 263 & 1 && 646 & 4 \\
        \cite{Chaurasiya2019} & 218 & 1 && 572 & 4 \\
        \cite{Jaiswal2019} & 273 & 1 && 682 & 4 \\
        \cite{Uguen2019} & 253 & 1 && 469 & 4 \\
        \cite{murillo2020customized} & 237 & 1 && 604 & 4 \\
        (prop.) & 185 & 0 && 435 & 0 \\
        \bottomrule
    \end{tabular}
\end{table}

\begin{figure*}[!t]
\centering
\subfloat[Area]{\includegraphics[width=0.32\textwidth]{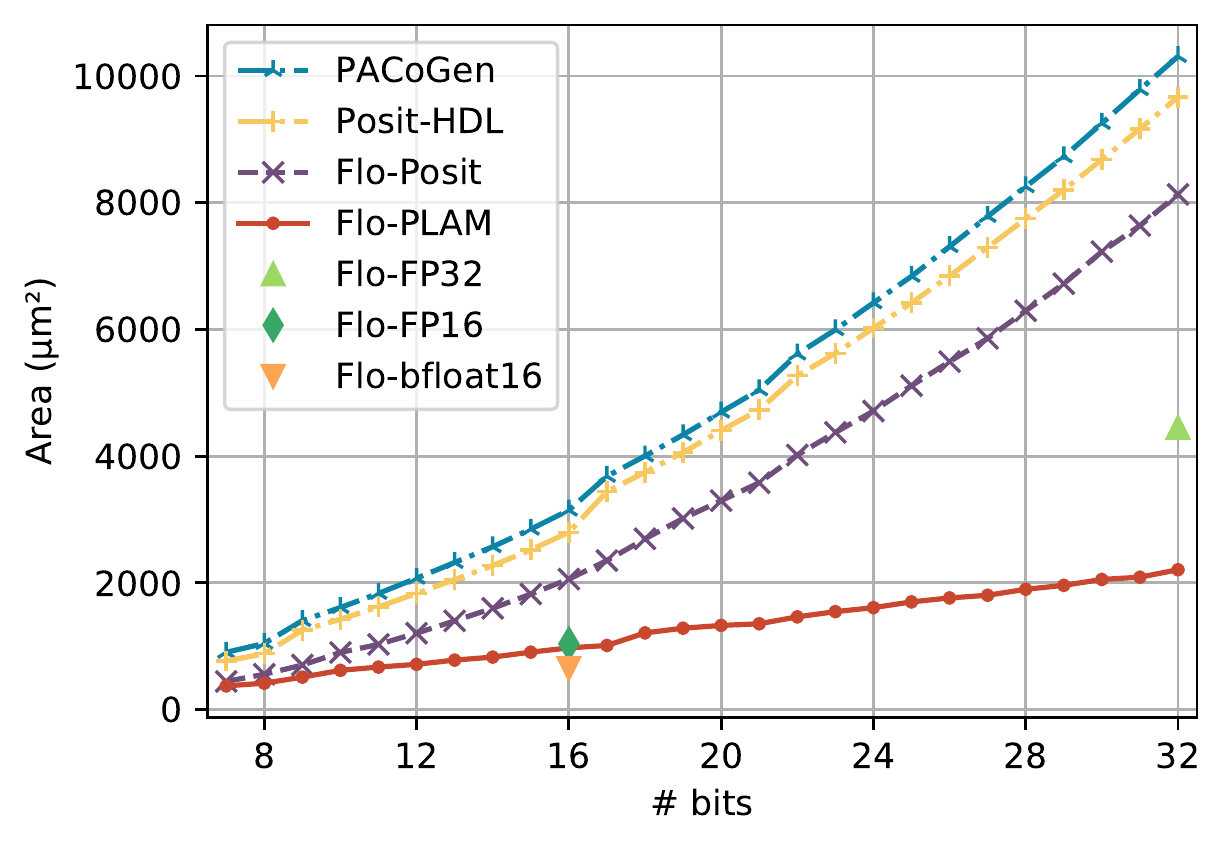}%
\label{fig:ASIC_area}
}
\hfil
\subfloat[Total delay]{\includegraphics[width=0.32\textwidth]{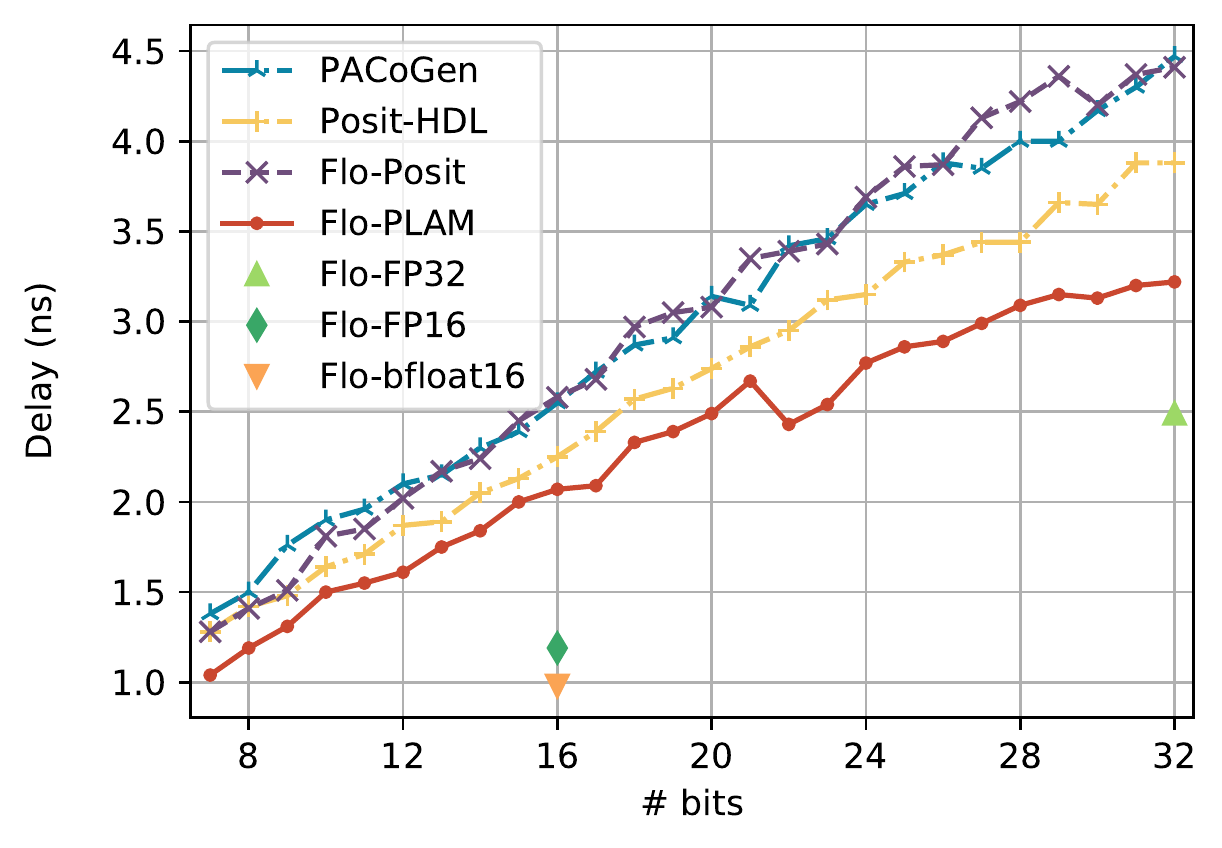}%
\label{fig:ASIC_delay}
}
\hfil
\subfloat[Power]{\includegraphics[width=0.32\textwidth]{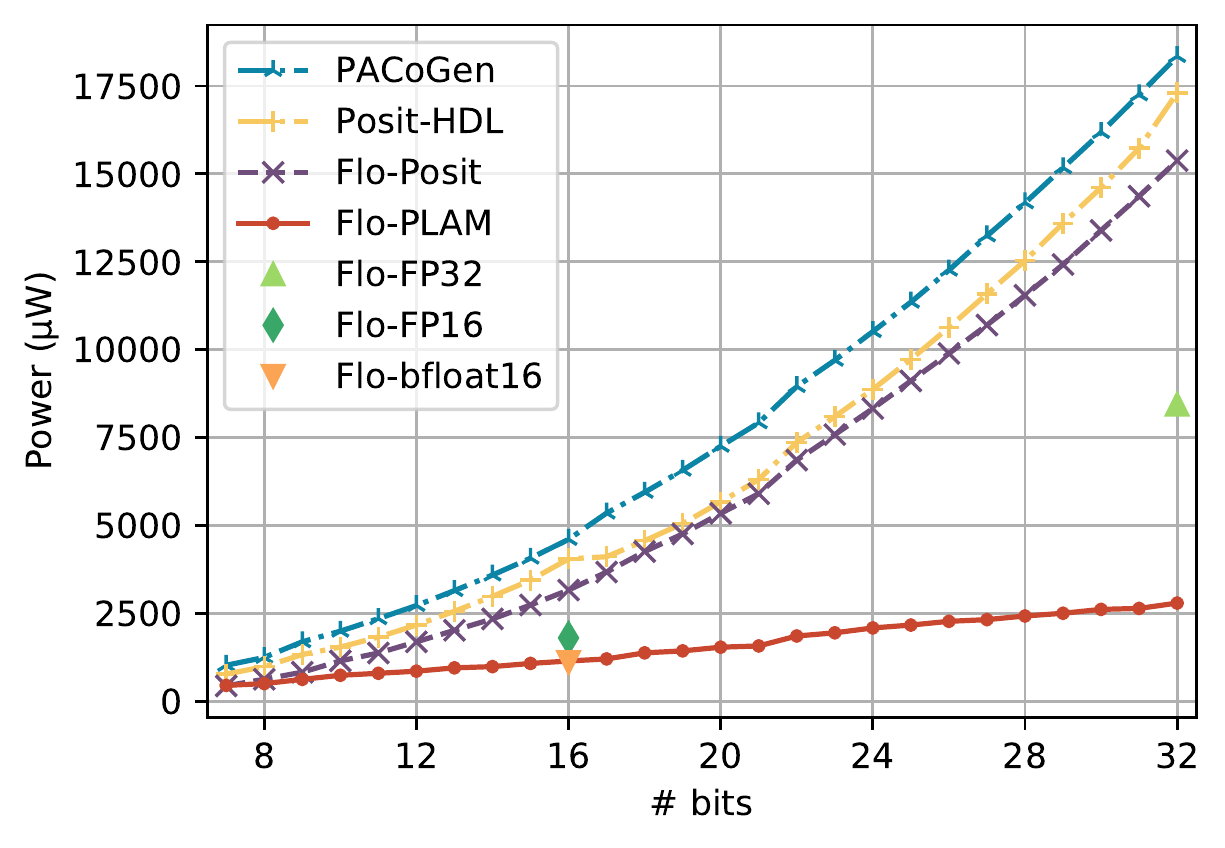}%
\label{fig:ASIC_power}
}
\caption{\positenv{n}{2} and floating-point multiplier implementation results. The prefix `Flo' stands for FloPoCo-generated.}
\label{fig:ASIC_cmp}
\end{figure*}

Standard cell synthesis has been performed using Synopsys Design Compiler with a 45-nm library by TSMC for the proposed generated units and open-source implementations.
The results for $es=2$ are compared graphically in Fig. \ref{fig:ASIC_cmp}. Similar results are obtained for different exponent sizes. As can be seen, area and power savings are greater as the bitwidth increases, obtaining respective reductions of $69.06\%$ and $63.63\%$ in the 16-bit case and of $72.86\%$ and $81.79\%$ for 32-bit multipliers in comparison with units from \cite{murillo2020customized}.
Besides, significant resource savings are obtained compared to the single-precision floating-point operator.
Under the same 32-bit length, the proposed posit approximate multiplier reduces area and power by $50.40\%$ and $66.86\%$, respectively.
These savings decrease in the 16-bit case, but the area and power usage is still lower than for the half-precision multiplier, closer to the bfloat16 one.
Furthermore, it must be reminded that these FloPoCo-generated units do not consider special cases, while the posit-based units do not have to do this.

On the other hand, the delay-reduction
is not as pronounced as in the previous case (up to a $17.01\%$ with respect to the 32-bit multiplier from Posit-HDL), and it is still higher than the corresponding floating-point operator under the same bitwidth.
This is due to the complexity of detecting the variable-length fields of posits, which is a design challenge for this format.

\begin{figure}[!t]
    \centering
    \includegraphics[width=\linewidth]{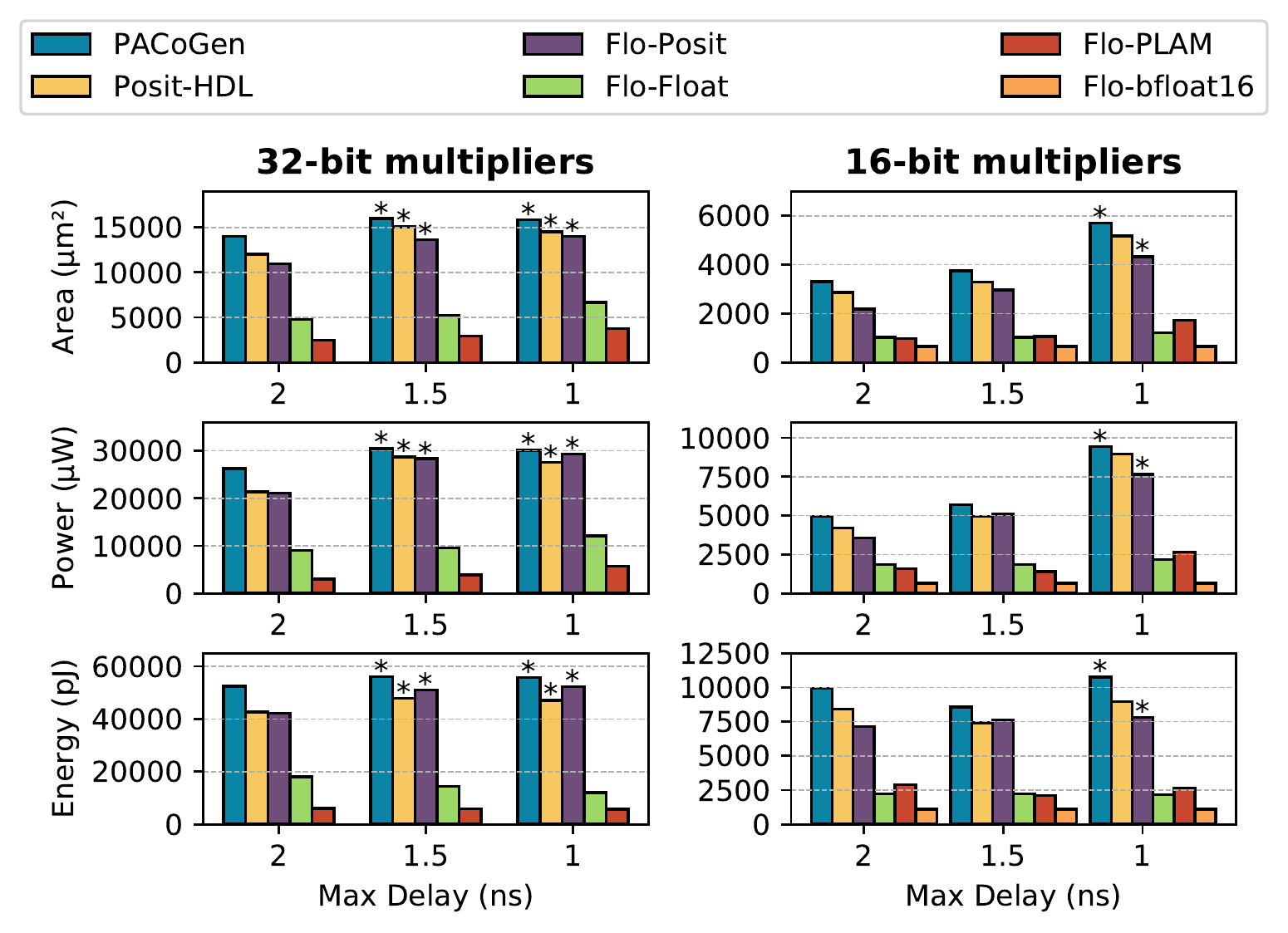}
    \caption{Results for time-constrained multiplier implementations. Implementations marked with `*' do violate the maximum delay constraint.}
    \label{fig:constrained_results}
\end{figure}

For a more thorough evaluation of the proposed design, area, power and energy are evaluated in different scenarios with a maximum delay constraint. As can be seen in the results depicted in Fig. \ref{fig:constrained_results}, the approximate 32-bit posit multiplier is by far more efficient than exact posit units, and even better than the equivalent floating-point unit, while in the case of 16-bits, the resource consumption of PLAM is similar to that produced by floating-point multipliers. Only the Flopoco-generated bfloat16 implementation shows better figures.

\section{Conclusions}
\label{sec:conclusions}

While the posit format has demonstrated to be a promising alternative
to the IEEE-754 floating-point standard
for DNNs, arithmetic units are still far from being competitive in terms of power.
This paper aims to reduce such a gap by proposing a Posit Logarithm-Approximate Multiplication (PLAM) scheme to reduce posit multiplication complexity. 
The experimental results show that applying PLAM in DNN inference does not affect accuracy.
When compared to other posit hardware solutions, the proposed implementation achieves area, power, and delay reduction of 72.86\%, 81.79\%, and 17.01\%, respectively.


\bibliographystyle{IEEEtran}
\bibliography{references.bib}

\end{document}